%% file: syscomNN.tex
\documentclass[11pt,a4paper]{article}
\usepackage{acl2015}
\usepackage{times}
\usepackage{url}
\usepackage{latexsym}
\usepackage{amsmath}
\usepackage{amssymb}
\usepackage{array}
\usepackage[ngerman,english]{babel}
\usepackage{breakcites}
\usepackage{calc}
\usepackage{color}
\usepackage{colortbl}
\usepackage[final]{graphicx}
\usepackage{indentfirst}
\usepackage[utf8]{inputenc}
\usepackage{import}
\usepackage{makerobust}
\usepackage{multicol}
\usepackage{multirow}
\usepackage{pgfplots}
\usepackage{pgfplotstable}
\usepackage{relsize}
\usepackage{stmaryrd}
\usepackage[normalsize]{subfigure}
\usepackage{varioref}
\usepackage{xspace}
\usepackage{xstring}
\usepackage{times}
\usepackage{url}
\usepackage{latexsym}
\usepackage{graphicx}
\usepackage{listings}
\usepackage{xcolor}
\usepackage{mathptmx}
\usepackage[nomessages]{fp}
\usepackage{amssymb,epsfig}
\usepackage{fancyvrb}
\usepackage{booktabs}
\usepackage{float}

\def\roundposition{1}

\edef\rounded{0}
\newcommand{\rdm}[1]{\edef\rounded{0}\FPeval\rounded{round(#1,\roundposition)}\rounded}

\newcommand\BLEU{\textsc{Bleu}\xspace}
\newcommand\sBLEU{\textsc{sBleu}\xspace}
\newcommand\TER{\textsc{Ter}\xspace}

\newcommand{\GIZA}{{GIZA\nolinebreak[4]\hspace{-.025em}\raisebox{.2ex}{\small\bf++}}\xspace}

\graphicspath{ {pictures/}{} }

\newenvironment{packed_descr}{
\begin{description}
  \setlength{\itemsep}{1pt}
  \setlength{\parskip}{-1pt}
  \setlength{\parsep}{0pt}
}{\end{description}}

\title{Local System Voting Feature for Machine Translation System Combination}

\author{Markus Freitag, Jan-Thorsten Peter, Stephan Peitz, Minwei Feng and Hermann Ney\\
        Human Language Technology and Pattern Recognition Group \\
        Computer Science Department \\
        RWTH Aachen University \\
        D-52056 Aachen, Germany \\
        {\tt <surname>@cs.rwth-aachen.de }}

\date{}

\begin{document}
\maketitle
\begin{abstract}

In this paper, we enhance the traditional confusion network system combination 
approach with an additional model trained by a neural network. 
This work is motivated by the fact that the commonly used binary system voting models
only assign each input system a global weight which is responsible for the global impact of
each input system on all translations. 
This prevents individual systems with low system weights from having influence on the system
combination output, although in some situations this could be helpful.
Further, words which have only been seen by one or few systems rarely
have a chance of being present in the combined output.
We train a local system voting model by a neural network which is based on
the words themselves and the combinatorial occurrences of the different system outputs.
This gives system combination the option to prefer other systems
at different word positions even for the same sentence.

\end{abstract}

\section{Introduction}
\label{sec:syscomNN:introduction}

Adding more linguistic informed models (e.g. language model or translation model) 
additionally to the standard models into system combination seems to yield no or only small improvements.
The reason is that all these models should have already been applied during the decoding 
process of the individual systems (which serve as input hypotheses for system combination)
and hence already fired before system combination. To improve system combination with additional models,
we need to define a model which can not be applied by an individual system.

In state-of-the-art confusion network system combination the following models are usually applied:

\begin{packed_descr}
\item[System voting (globalVote) models]
  For~each word the voting model for system $i$ ($1 \leq i \leq I$) is~1 iff the
  word is from system $i$, otherwise~0.
\item[Binary primary system model (primary)]
  A~model that marks the primary hypothesis.
\item[Language model]
  3-gram language model (LM) trained on the input hypotheses.
\item[Word penalty]
  Counts the number of words.
\end{packed_descr}

To gain improvements with additional models, it is better to define models
which are not used by an individual system. A simple model which can not be applied by any individual system
is the binary system voting model (globalVote). This model is the most important one during
system combination decoding as it determines the impact of each individual system.
Each system $i$ is assigned one globalVote model which fires if the word is generated by system $i$.
Nevertheless, this simple model is independent of the actual words and the score is only based
on the global preferences of the individual systems. This disadvantage prevents system
combination from producing words which have only been seen by systems
with low system weights (low globalVote model weights). To give systems and words
with low weights a chance to affect the final output, we define
a new local system voting model (localVote) which makes decisions based on the current word options and not only
on a general weight. The local system voting model allows system combination to prefer different system outputs
at different word positions even for the same sentence.

Motivated by the success of neural networks in language modelling \cite{bengio2006neural,schwenk2002connectionist}
and translation modelling \cite{Son:2012:CST:2382029.2382036}, we choose 
feedforward neural networks to train the novel model. Instead of
calculating the probabilities in a discrete space, the neural network projects the
words into a continuous space. This projection gives us the option to
assign probability also to input sequences which were not observed in the training data.
In system combination each training sentence has to be translated by all individual 
system engines which is time consuming. Due to this we have
a small amount of training data and thus it is very likely
that many input sequences of a test set have not be seen during training.

The remainder of this paper is structured as follows:
in Section~\ref{sec:syscomNN:relatedWork}, we discuss some related work.
In Section~\ref{sec:syscomNN:newvmodel}, the novel local system voting model is described.
In Section~\ref{sec:syscomNN:experiments}, experimental results are 
presented which are analyzed in Section~\ref{sec:syscomNN:analysis}.
The paper is concluded in Section~\ref{sec:syscomNN:conclusion}.

\section{Related Work}
\label{sec:syscomNN:relatedWork}

In confusion network decoding, pairwise alignments between all system outputs are generated.
From the calculated alignment information, a confusion network is built from which the 
system combination output is determined using majority voting and additional models.
The hypothesis alignment algorithm is a crucial part of building the confusion network and many alternatives have been proposed in the literature:

    \begin{packed_descr}

    \item[\cite{bangalore01computing}] use a multiple string alignment (MSA) algorithm to identify the unit of consensus and applied a posterior language
    model to extract the consensus translations. In contrast to the following approaches, MSA is unable to capture word reorderings.

    \item[\cite{matusov06computing}] produce pairwise word alignments with the statistical alignment algorithm toolkit \GIZA that explicitly models word reordering.
    The context of a whole document of translations rather than a single sentence is taken into account to
    produce the alignments.

    \item[\cite{sim07consensus}] construct a consensus network by using \TER~\cite{snover06study} alignments.
    Minimum bayes risk decoding is applied to obtain a primary hypothesis to which all other hypotheses are aligned.

    \item[\cite{rosti2007combining}] extend the \TER alignment approach and introduce an incremental \TER alignment which
    aligns one system at a time to all previously aligned hypotheses.

    \item[\cite{karakos2008machine}] use the inversion transduction grammar (ITG) formalism \cite{wu1997stochastic} and 
    treat the alignment problem as a problem of bilingual parsing to generate the pairwise alignments. 

    \item[\cite{he2008indirect}] propose an indirect hidden markov model (IHMM) alignment approach to address the synonym matching
    and word ordering issues in hypothesis alignment.

    \item[\cite{Heafield-marathon}] use the METEOR toolkit to calculate pairwise alignments between the hypotheses. 

    \end{packed_descr}

All confusion network system combination approaches only
use the global system voting models. Regarding to this chapter, there has been similar effort
in the area of speech recognition:

\begin{packed_descr}

    \item[\cite{hillard2007rover}] Similar work has been presented for system combination
    of speech recognitions systems: the authors train a classifier to learn which 
    system should be selected for each output word. The learning target for each slot
    is the set of systems which match the reference word, or the null class if no systems 
    match the reference word. Their novel approach outperforms the ROVER baseline by up to 14.5\% relatively on an evaluation set.

\end{packed_descr}

\section{Novel Local System Voting Model}
\label{sec:syscomNN:newvmodel}
In the following subsections we introduce a novel local system voting model (localVote) trained by a neural
network. The purpose of this model is to prefer one particular path in the 
confusion network and therefore all local word decisions between two nodes leading
to this particular path.
More precisely, we want the neural network to learn an oracle path extracted 
from the confusion network graph which leads to the lowest error score. 
In Subsection~\ref{subsec:syscomNN:bestPath}, we describe a polynomial
approximation algorithm to extract the best sentence level \BLEU (\sBLEU)
path in a confusion network.
Taking this path as reference path, we define the model in Subsection~\ref{subsec:syscomNN:trainNN}
followed by its integration in the linear model combination in Subsection~\ref{subsec:syscomNN:scoreNN}.

\subsection{Finding \sBLEU-optimal Hypotheses}
\label{subsec:syscomNN:bestPath}

In this section, we describe a polynomial approximation
algorithm to extract the best \sBLEU hypothesis from a confusion network.
\cite{leusch08:cnbleu} showed that this problem is generally
NP-hard for the popular \BLEU~\cite{papineni2002bleu} metric.
Nevertheless, we need some paths which serve as ``reference paths``.

Using \BLEU as metric to extract the best possible path is
problematic as in the original \BLEU definition there is no
smoothing for the geometric mean. This has the
disadvantage that the \BLEU score becomes zero already
if the four-gram precision is zero, which 
can happen obviously very often with short or difficult translations.
To allow for sentence-wise evaluation, we use
the \sBLEU metric \cite{lin2004automatic}, which is basically \BLEU
where all $n$-gram counts are initialized
with 1 instead of 0. The brevity penalty is calculated only
on the current hypothesis and reference sentence.

We use the advantage that confusion networks can be sorted topologically.
We walk the confusion network from the start node to the end node, keeping track
of all $n$-grams seen so far.
At each node we keep a $k$-best list containing the partial
hypotheses with the most $n$-gram matches leading to this node and recombine only
partial hypotheses containing the same translation.
As the search space can become exponentially large, we only keep
$k$ possible options at each node. This pruning can lead to
search errors and hence yield non-optimal results. If needed for hypotheses
with the same $n$-gram counts, we prefer hypotheses with a higher translation
score based on the original models.
For the final node we add the brevity penalty
to all possible translations. 

\begin{figure}[ht!]
\centering
\def\svgwidth{0.9\linewidth}
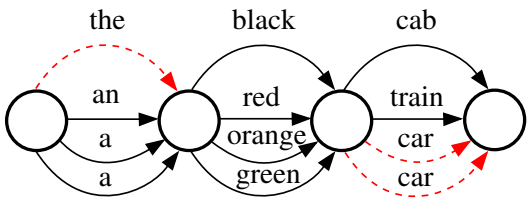
\caption{System A: \emph{the black cab} ; System B: \emph{an red train} ; System C: \emph{a orange car} ;  System D: \emph{a green car} ; Reference: \emph{the blue car} .}
\label{fig:4}
\end{figure}

\begin{figure}[ht!]
\centering
\def\svgwidth{0.9\linewidth}
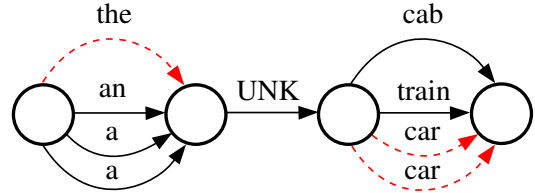
\caption{As the words \emph{black}, \emph{red}, \emph{orange}, and \emph{green} in Figure~\ref{fig:4} are all not present in the reference (\emph{the blue car}), they are mapped to one single ''UNK'' arc.}
\label{fig:4b}
\end{figure}

As we are only interested in arc decisions which match a reference word,
we simplify the confusion network before applying the algorithm. 
If all arcs between two adjacent nodes are not present in the reference, we remove all of them and
add a single arc labeled with ''UNK''.
This reduces the vocabulary size and still gives us the same best \sBLEU scores as before.
In Figure~\ref{fig:4}, a confusion network of four input hypotheses is given.
As the words \emph{black}, \emph{red}, \emph{orange}, and \emph{green} are all not present in the
reference, all of them are mapped to one single ''UNK'' arc (cf. Figure~\ref{fig:4b}). The best \sBLEU
path is \emph{the UNK car}.

\subsection{localVote Model Training}
\label{subsec:syscomNN:trainNN}

The purpose of the new localVote model is to prefer the best \sBLEU path and therefore to learn
the word decisions between all adjacent nodes which lead to this particular path.
During the extraction of the best \sBLEU hypotheses from the confusion network, we keep track of all arc decisions.
This gives us the possibility to generate local training examples based only on the $I$ arcs
between two nodes.
For the confusion network illustrated in Figure \ref{fig:4b}, we generate two training examples for the neural network training. Based
on the arcs \emph{the}, \emph{an}, \emph{a} and \emph{a} we learn the output \emph{the}.
Based on the arcs \emph{cab}, \emph{train}, \emph{car} and \emph{car} we learn the output \emph{car}.

In all upcoming system setups, we use the open source toolkit NPLM~\cite{vaswanidecoding} for
training and testing the neural network models. We use the standard setup as described in the
paper and use the neural network with one projection layer and one hidden layer. For
more details we refer the reader to the original paper of the NPLM toolkit.
The inputs to the neural network are the $I$ words produced by the $I$ different
individual systems. The outputs are the posterior probabilities of all words of the vocabulary.
The input uses the so-called 1-of-$n$ coding, i.e. the $i$-th word of the vocabulary is coded by setting
the $i$-th element of the vector to 1 and all the other elements to 0.

\begin{figure}[t!]
\centering
\def\svgwidth{\linewidth}
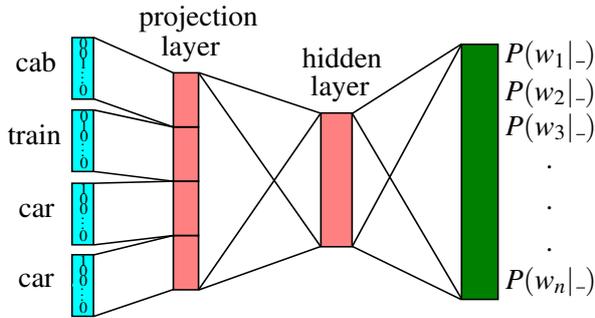
\caption{Unigram neural network training example: System $A$ produces \emph{cab}, System $B$ \emph{train}, System $C$ \emph{car}, System $D$ \emph{car}, reference is \emph{car}. 1-of-$n$ encoding was applied to map words to a suitable neural network input.}
\label{fig:trainingExample}
\end{figure}

For a system combination of $I$ individual systems, a training example consists of $I+1$ words. The first $I$ words
(input of the neural network) are representing the words of the individual systems, the last position (output of the neural network) serves as slot for the
decision we want to learn (extracted from the best \sBLEU path). We do not add the ''UNK'' arcs to the neural network training as they do not help to increase the \sBLEU score.
Figure~\ref{fig:trainingExample} shows the neural network training example for the last words of Figure~\ref{fig:4b}.
The output of each individual system provides one input word. In Table~\ref{tab:unigram-trainingExamples} the
two training examples for Figure~\ref{fig:4b} are illustrated. 

As a neural network training example only consists of the $I$ words between two adjacent nodes, we are able to produce several training examples for each sentences.
For a system combination of $I$ systems and a development set of $S$ sentences with an average sentence length of $L$, we can
generate up to $I*S*L$ neural network training examples.

\begin{table}[t!]
\caption{Training examples from Figure~\ref{fig:4b}.}
\begin{center}
\begin{tabular}{|lccc|c|}
\hline
    \multicolumn{4}{|c|}{input layer} &  \\
    Sys A & Sys B & Sys C & Sys D & ref \\ \hline
     the & an & a & a & the \\
     cab & train & car & car & car \\ \hline
\end{tabular}
\label{tab:unigram-trainingExamples}
\end{center}
\end{table}

\begin{figure}[t!]
\centering
\def\svgwidth{\linewidth}
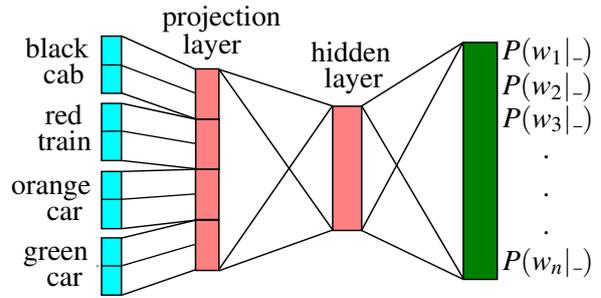
\caption{Bigram neural network training example: System $A$ produces \emph{black cab}, System $B$ \emph{red train}, System $C$ \emph{orange car}, System D \emph{green car}, reference is \emph{car}. }
\label{fig:trainingExample2}
\end{figure}

Further, we can expand the model to use arbitrary history size, if we take the predecessor words into account.
Instead of just using the local word decision of a system, we add additionally the predecessors
of the individual systems into the training data.
In Figure~\ref{fig:trainingExample2}, we e.g. use the bigram \emph{red train} instead of the unigram
\emph{train} for system $B$ into the training data. In Table~\ref{tab:unigram-trainingExamples-bigram} all bigram
training examples of Figure~\ref{fig:4b} can be seen.

\subsection{localVote model Integration}
\label{subsec:syscomNN:scoreNN}

Having a trained localVote model, we then add it as an additional model into the confusion network.
We calculate for each arc the probability of the word in the trained neural network.
E.g. for Figure~\ref{fig:4}, we extract the probabilities for all arcs by the strings
illustrated in Table~\ref{tab:unigram-scoringSequences}.
Finally, we add the scores as a new model
and assign it a weight which is trained
additionally to the standard model weights with MERT.

\begin{table}[t]
\caption{Training examples (bigram) from Fig.~\ref{fig:4b}.}
\begin{center}
\setlength{\tabcolsep}{0.35em}
\begin{tabular}{|lccc|c|}
\hline
    \multicolumn{4}{|c|}{input layer} &  \\
    Sys $A$ & Sys $B$ & Sys $C$ & Sys $D$ & ref \\ \hline
    \textless s\textgreater   the & \textless s\textgreater   an & \textless s\textgreater   a & \textless s\textgreater   a & the \\
    black cab & red train & orange car & green car & car \\ \hline
\end{tabular}
\label{tab:unigram-trainingExamples-bigram}
\end{center}
\end{table}

\begin{table}[h]
\caption{Calculating the probability for all possible output words from Figure~\ref{fig:4}. The output layer is the current generated  word.}
\begin{center}
\begin{tabular}{|lccc|c|}
\hline
    \multicolumn{4}{|c|}{input layer} &  \\
    Sys A & Sys B & Sys C & Sys D & arc word \\ \hline
the & an & a & a & the \\
the & an & a & a & an \\
the & an & a & a & a \\
black & red & orange & green & black \\
black & red & orange & green & red \\
black & red & orange & green & orange \\
black & red & orange & green & green \\
cab & train & car & car & cab \\
cab & train & car & car & train \\
cab & train & car & car & car \\ \hline
\end{tabular}
\label{tab:unigram-scoringSequences}
\end{center}
\end{table}

\subsection{Word Classes}
\label{subsec:syscomNN:wordclasses}

The neural network training sets are relatively small as all sentences have to be translated by
all individual system engines. This results in many
unseen words in the test sets. To overcome this problem, we use word classes~\cite{mkcls} instead of words 
which were trained (10 iterations) on the target part of the bilingual training corpus in some experiments. We use
the trained word classes on both input layer and output layer.

\section{Experiments}
\label{sec:syscomNN:experiments}

All experiments have been conducted with the open source system combination toolkit Jane~\cite{freitag14:jane}.
For training and scoring neural networks, we use the open source toolkit NPLM~\cite{vaswanidecoding}.
NPLM is a toolkit for training and using feedforward neural language models. 
Variations in neural network architecture have been tested. We tried various hidden layer sizes 
as well as projection layer sizes. We achieved similar results for all setups and decided
to stick to 1 hidden layer whose size is 200, a learning rate of 0.08 and let the training run
20 epochs in all experiments.

Translation quality is measured in lowercase with \BLEU~\cite{papineni2002bleu} and \TER~\cite{snover06study}
whereas the performance of each setup is the best score on the tune set across five different MERT runs.
The system combination weights of the linear model are optimized with MERT on 200-best lists
with (\TER-\BLEU)/2 as optimization criterion. For all language pairs we use three different test sets. In the following the test set
for extracting the training examples for the neural network training is labeled as \emph{tune (NN)}.
The test set \emph{tune (MERT)} indicates the tune set for MERT and \emph{test}
indicates the blind test set.

The individual systems are different extensions of phrase-based or hierarchical phrase-based systems.
The systems are built on the same amount of preprocessed training data and differ mostly in the
models which are used to score the translation options. Further, some systems are syntactical
augmented based on syntax trees on either source or target side.

\subsection{BOLT Chinese$\to$English}
\label{subsec:syscomNN:experiments-chen}

For Chinese$\to$English, we use the current BOLT data set (corpus statistics 
are given in Table~\ref{tab:corpora:Stats-ChEn}).
The test sets consist of text drawn from ''discussion forums'' in Mandarin Chinese.
We use nine individual systems to perform the system combination experiments.
The lambda weights are optimized on a tune set of 985 sentences (tune (MERT)). 
We train the proposed localVote model on 15,323,897 training examples extracted from the
1844 sentences tune (NN) set.

\begin{figure}[ht!]
\centering
\def\svgwidth{\linewidth}
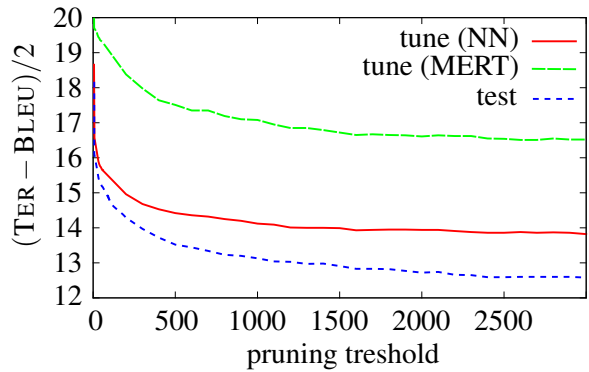
\caption{$(\TER - \BLEU)/2$ scores for different $k$-best pruning thresholds on the BOLT Chinese$\to$English data set.}
\label{fig:ChEn.BleuThreshold}
\end{figure}

As a first step we have to determine the $k$-best pruning threshold for extracting
the \sBLEU optimal path from the current confusion networks (cf. Section~\ref{subsec:syscomNN:bestPath}).
In Figure~\ref{fig:ChEn.BleuThreshold} the (\TER-\BLEU)/2 results
of the \sBLEU optimal hypotheses extracted with different $k$-best sizes are given.
Although, the \BLEU score improves by setting $k$ to a higher value,
the computational time increases.
To find a tradeoff between running time and performance, we set the $k$-best size to 1200
in the following experiments.

\begin{table}[ht!]
\caption{Corpus statistics Chinese$\to$English.}
\begin{center}
\begin{tabular}{|l|c|c|}
\hline 
             &  Chinese    &   English      \\ \hline
Sentences    & \multicolumn{2}{c|}{13M}\\ \hline
Running words& 255M & 279M \\ \hline
Vocabulary   & 370K & 833K \\ \hline
Tune sentences & \multicolumn{2}{c|}{1844 (NN), 985 (MERT)}\\ \hline
Test sentences  & \multicolumn{2}{c|}{1124}\\ \hline
\end{tabular}
\label{tab:corpora:Stats-ChEn}
\end{center}
\end{table}

Experimental results are given in Table~\ref{tab:ChEnResults}.
The \emph{baseline} is a system combination run without 
any localVote model of nine individual systems using the
standard models as described in \cite{freitag14:jane}. 
The \emph{oracle} score is calculated on the hypothesis
of the \sBLEU best path extracted with $k=1200$. 
We train the neural network on 15,323,897 training examples generated from the 1844 tune (NN) sentences. 
By training a neural network based on unigram decisions (\emph{unigram NN}), we gain
small improvements of -0.6 points in \TER.
As we have only few sentences of training data, many words have
not been seen during neural network training. To overcome this problem, we train
1500 word classes on the target part of the bilingual data.
Learning the localVote model on word classes (\emph{unigram wcNN}) gain
improvement of +0.7 points in \BLEU and -0.6 points in \TER. 
By taking a bigram history
into the training of the neural network, we reach only small further improvement.
Compared to the \emph{baseline}, the system combination \emph{+bigram NN} 
outperforms the \emph{baseline} by +0.3 points in \BLEU and -0.6 points in \TER.
By using word classes (\emph{+bigram wcNN}) we gain improvement of +0.4 points in \BLEU and -1.0 points
in \TER. 

\begin{table}[t!]
      \caption{Results for the BOLT Chinese$\to$English translation task. The localVote models of the systems \emph{+unigram~NN} and \emph{+unigram~wcNN} are trained based on one word per system.
  The localVote models of the systems \emph{+bigram~NN} and \emph{+bigram~wcNN} are trained based on two words per system. For systems labeled with \emph{wcNN}, the neural network is trained on word classes.
Significance is marked with $\dagger$ for 95\% confidence and $\ddagger$ for 99\% confidence, and is measured with the bootstrap resampling method as described in \cite{koehn2004statistical}.}
  \begin{center}
\setlength{\tabcolsep}{0.36em}
    \begin{tabular}{|l|l|l|l|l|}
	\hline
      \bf{system} & \multicolumn{2}{c|}{\bf{tune}} & \multicolumn{2}{c|}{\bf{test}} \\
      & \BLEU & \TER & \BLEU & \TER \\
       \hline \hline
        \textbf{baseline}                  & \rdm{17.89} & \rdm{61.45} & \rdm{18.3} & \rdm{60.9} \\ \hline 
        \textbf{\emph{+unigram NN}}     & \rdm{18.13} & \rdm{61.24} & \rdm{18.34} & \rdm{60.26}$\dagger$ \\ 
        \textbf{\emph{+unigram wcNN}}     & \rdm{18.41} & \rdm{61.52} & \rdm{18.98}$\ddagger$ & \rdm{60.28}$\dagger$ \\ 
        \textbf{\emph{+bigram NN}}      & \rdm{18.14} & \rdm{61.34} & \rdm{18.60}$\dagger$ & \rdm{60.33}$\dagger$ \\  
        \textbf{\emph{+bigram wcNN}}      & \rdm{18.13} & \rdm{61.19} & \rdm{18.72}$\dagger$ & \rdm{59.86}$\ddagger$ \\ \hline 
        \textbf{oracle}                  & \rdm{28.57} & \rdm{62.27} & \rdm{31.11} & \rdm{57.18} \\ \hline 
      \end{tabular}
      \label{tab:ChEnResults}
     \end{center}
   \end{table}

\begin{figure}[ht!]
\centering
\def\svgwidth{\linewidth}
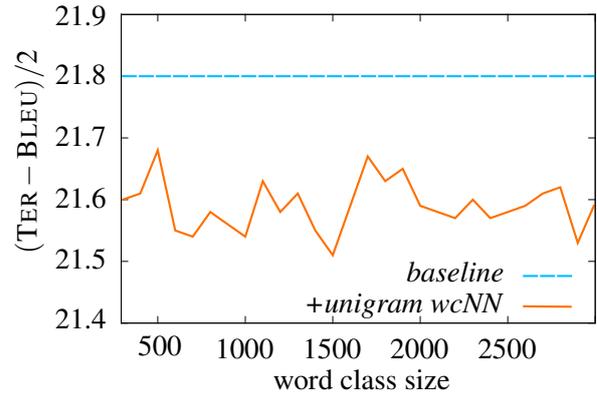
\caption{$(\TER - \BLEU)/2$ scores for different word class sizes on the BOLT Chinese$\to$English tune (MERT) set.}
\label{fig:ChEnWcResults}
\end{figure}

All results are reached with a word class size of 1500.
In Figure~\ref{fig:ChEnWcResults} the $(\TER - \BLEU)/2$ scores on tune(MERT) of
system combinations including one unigram localVote model trained with different
word class sizes are illustrated. Independent of the word class size,
system combination including a localVote model always performs better compared to the
baseline. The best performance is reached by a word class size of
1500. 
One reason for the loss of performance when using no word classes is the size of the neural network tune set.
Within a size of 1844 sentences, many words of the test set have never been
seen during neural network training. The test set has a vocabulary size of 6106
within 2487 words (40.73\%) are not present in the training set (tune (NN)) of the neural network.
For the MERT tune set 2556 words (40.91\%) are not present in the neural network training set.
Word classes tackle this problem and it is much more likely that each word class has been seen during the training procedure of the neural network.

\subsection{BOLT Arabic$\to$English}
\label{subsec:syscomNN:experiments-aren}

For Arabic$\to$English, we use the current BOLT data set (corpus statistics 
are given in Table~\ref{tab:corpora:Stats-ArEn}). The test sets consist of text
drawn from ''discussion forums'' in Egyptian Arabic.
We train the neural network on 6,591,158 training examples extracted from the 1510 sentences tune (NN) dev set.
The model weights are optimized on a 1080 sentences tune set.
All results are system combinations of five individual systems. The test set has a vocabulary size of 3491
within 1510 words (43.25\%) are not present in the training set (tune (NN)) of the neural network.
For the MERT tune set 1549 words (43.24\%) are not part of the neural network training set.

\begin{figure}[ht!]
\centering
\def\svgwidth{\linewidth}
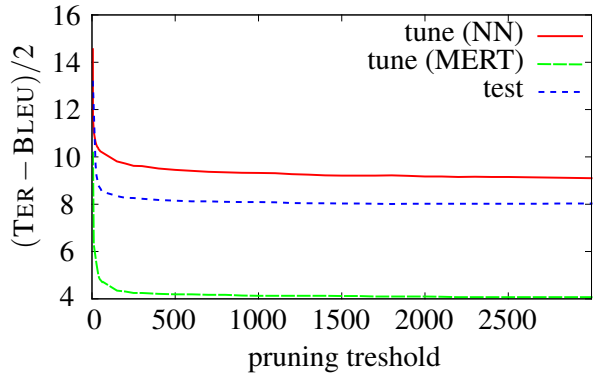
\caption{$(\TER - \BLEU)/2$ scores for different $k$-best pruning thresholds on the BOLT Arabic$\to$English tune (MERT) set.}
\label{fig:ArEn.BleuThreshold}
\end{figure}

We run the same experiment pipeline as for Chinese$\to$English and first
determine the $k$-best threshold for getting the oracle paths in the confusion networks.
As the Arabic$\to$English system combination is only based on 5 individual
systems, the confusion networks are much smaller. We set the pruning threshold to 1000
($k=1000$) which is a good tradeoff between running time and performance. Figure~\ref{fig:ArEn.BleuThreshold}
shows the $(\TER - \BLEU)/2$ scores for different $k$-best pruning thresholds. Increasing
$k$ to a higher value then 1000 improves the $(\TER - \BLEU)/2$ only slightly.

\begin{table}[ht!]
\caption{Corpus statistics BOLT Arabic$\to$English.}
\begin{center}
\begin{tabular}{|l|c|c|}
\hline 
             &  Arabic    &   English      \\ \hline
Sentences    & \multicolumn{2}{c|}{8M}\\ \hline
Running words& 189M & 186M \\ \hline
Vocabulary   & 608K & 519K \\ \hline
Tune sentences & \multicolumn{2}{c|}{1510 (NN), 1080 (MERT)}\\ \hline
Test sentences  & \multicolumn{2}{c|}{1137}\\ \hline
\end{tabular}
\label{tab:corpora:Stats-ArEn}
\end{center}
\end{table}

Experimental results are given in Table~\ref{tab:ArEnResults}.
The \emph{baseline} is a system combination run without 
any localVote model of five individual systems using the
standard models as described in \cite{freitag14:jane}. 
The \emph{oracle} score represents
the score of the \sBLEU best path extracted with $k=1000$. 
Training a localVote model based on the best \sBLEU path (\emph{+unigram NN}) gives
us improvement of +0.9 points in \BLEU compared to the \emph{baseline}.
Adding bigram context to the neural network training (\emph{+bigram NN}) yields improvement of
+0.8 points in \BLEU compared to the \emph{baseline} system combination.
By training word classes on the bilingual part of the training data,
we gain additional improvements. When using word classes and a history
size of two, \emph{+bigram wcNN} yields the best performance with +1.1
points in \BLEU compared to the \emph{baseline}.

\begin{figure}[ht!]
\centering
\def\svgwidth{\linewidth}
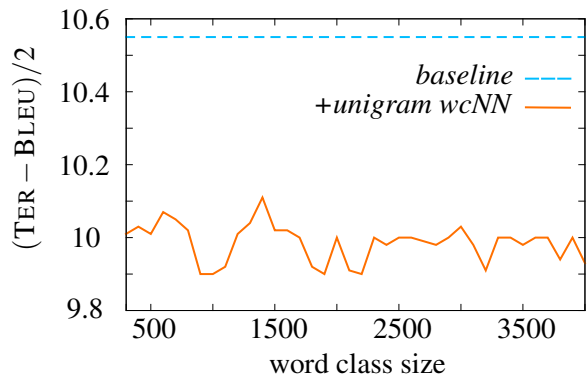
\caption{$(\TER - \BLEU)/2$ tune set scores for different word class sizes on the BOLT Arabic$\to$English task.}    
\label{fig:ArEnWcResults}
\end{figure}

All results are conducted with a word class size of 1000. The tune set performance
of different unigram localVote models trained on different word class sizes
are illustrated in Figure~\ref{fig:ArEnWcResults}.
The results are fluctuating and we set the word class size to 1000 in all Arabic$\to$English experiments.

\begin{table}[ht!]
      \caption{Results for the BOLT Arabic$\to$English translation task. The localVote models of the systems \emph{+unigram NN} and \emph{+unigram wcNN} are trained by a neural network based on one word per system.
  The localVote models of the systems \emph{+bigram NN} and \emph{+bigram wcNN} are trained by a neural network based on two words per system. For systems labeled with \emph{wcNN}, the neural network is trained on word classes for both input and output layer.
Significance is marked with $\ddagger$ for 99\% confidence and is measured with the bootstrap resampling method as described in \cite{koehn2004statistical}.}
    \begin{center}
    \setlength{\tabcolsep}{0.4em}
    \begin{tabular}{|l|l|l|l|l|}
      \hline
      \bf{system} & \multicolumn{2}{c|}{\bf{tune}} & \multicolumn{2}{c|}{\bf{test}} \\
      & \BLEU & \TER & \BLEU & \TER \\
       \hline \hline
       \textbf{baseline}                  & \rdm{30.13} & \rdm{51.21} & \rdm{27.6} & \rdm{55.8} \\ \hline 
          \textbf{\emph{+unigram NN}}     & \rdm{31.38} & \rdm{51.22} & \rdm{28.46}$\ddagger$ & \rdm{55.99} \\ 
        \textbf{\emph{+unigram wcNN}}     & \rdm{31.13} & \rdm{51.10} & \rdm{28.27}$\ddagger$ & \rdm{55.68} \\ 
         \textbf{\emph{+bigram NN}}      & \rdm{31.25} & \rdm{51.12} & \rdm{28.36}$\ddagger$ & \rdm{55.78} \\ 
       \textbf{\emph{+bigram wcNN}}      & \rdm{31.36} & \rdm{51.21} & \rdm{28.67}$\ddagger$ & \rdm{56.02} \\ \hline 
        \textbf{oracle}                  & \rdm{38.10} & \rdm{46.34} & \rdm{34.79} & \rdm{50.86} \\ \hline 
      \end{tabular}
      \label{tab:ArEnResults}
     \end{center}
   \end{table}

\section{Analysis}
\label{sec:syscomNN:analysis}

In this section we compare the final translations of the Chinese$\to$English
system combination \emph{+bigram wcNN} with the \emph{baseline}.
The word occurrence distributions for both setups are illustrated in Table~\ref{tab:Distri-ChEn}.
This table shows how many input systems produce a certain word and finally if it is part of the system combination output.
As the original idea of system combination is based on majority voting, it should be more likely that a word which is produced
by more input systems is in the final system combination output than a word which is only produced by few input systems.
E.g. 11008 words have been produced by all 9 individual systems from which all of them are in both the system combination \emph{baseline} and the advanced
system \emph{+bigram wcNN}. If a word is only produced by 8 individual systems, a ninth system does not produce this word.
98,9\% of the words produced by only 8 different individual systems are
in the final \emph{baseline} system combination output. The missing words result mostly from alignment errors produced by
the pairwise alignment algorithm when aligning the single systems together. 

We observe the problem that the globalVote
models prevent words, which have only been produced by few systems, to be present in the system combination output. 
In Table~\ref{tab:Distri-ChEn}, you can see that words which are only produced by 1-4 individual systems are more likely
to be present in the final output when including the novel localVote model. As e.g. in the baseline 592 of the 6129 words which
have only been produced by two individual system are in the output, the advanced \emph{+bigram wcNN} setup contains additional 172 words.
These statistics demonstrate the functionality of the novel localVote model which does not only improve the translation quality in terms of \BLEU,
but also tackles the problem of the dominating globalVote models.

\begin{table}[t!]
\caption{Word occurrence distribution for the Chinese$\to$English setup. First column indicates in how many systems a word appears. E.g. 120/14072 (0.9\%) indicates that 14072 words only appear in one individual input system from which 120 (0.9\%) are present in the baseline system combination hypothesis.}
\begin{center}
\setlength{\tabcolsep}{0mm}
\begin{tabular}{|c|rrr|rrr|}
\hline
   \hspace{0.005em}  \# \hspace{0.005em} &  \multicolumn{3}{c|}{ \emph{baseline} \hspace{0.02em}} & \multicolumn{3}{c|}{ \emph{+bigram wcNN} }\\ \hline
                     1 & \hspace{0.04em} 120/  & 14072 & (0.9\%) \hspace{0.07em}  & 214/ & 14072  & (1.5\%)  \\
                     2 & \hspace{0.04em} 592/  & 6129   & (9.7\%) \hspace{0.07em} & 764/ & 6129  & (12.5\%)  \\
                     3 & \hspace{0.04em} 1141/ & 4159  & (27.4\%) \hspace{0.07em} & 1319/ & 4159  & (31.7\%)  \\
                     4 & \hspace{0.04em} 1573/ & 3241 & (48.5\%) \hspace{0.07em} & 1669/ & 3241  & (51.5\%)  \\
                     5 & \hspace{0.04em} 2051/ & 2881  & (71.2\%) \hspace{0.07em} & 1993/ & 2881 & (69.2\%)  \\
                     6 & \hspace{0.04em} 2381/ & 2744  & (86.8\%) \hspace{0.07em} & 2332/ & 2744  & (85.0\%)  \\
                     7 & \hspace{0.04em} 2817/ & 2965  & (95.0\%) \hspace{0.07em} & 2820/ & 2965  & (95.1\%)  \\
                     8 & \hspace{0.04em} 3818/ & 3860  & (98.9\%) \hspace{0.07em} & 3815/ & 3860  & (98.8\%)  \\
                     9 & \hspace{0.04em} 11008/ & 11008  & (100.0\%) \hspace{0.07em} & \hspace{0.02em} 11008/ & 11008 & (100.0\%) \\ \hline
\end{tabular}
\label{tab:Distri-ChEn}
\end{center}
\end{table}

The Arabic$\to$English word occurrence distribution is illustrated in Table~\ref{tab:Distri-ArEn}. A similar scenario as
for the Chinese$\to$English translation task can be observed. The words which only occur in few individual systems
have a much higher chance to be in the final output when using the novel local voting system model. It is also visible 
that the neural network model prevents some words of being in the combined output even if the word have been produced
by 4 of 5 systems. The novel local system voting model gives system combination the option to select words which have 
only be generated by few individual systems.

\begin{table}[ht!]
\caption{Word occurrence distribution for the Arabic$\to$English setup. First column indicates in how many systems a word appears. E.g. 214/5791 (3.7\%) indicates that 5791 words only appear in one individual input system from which 214 (3.7\%) are present in the baseline system combination hypothesis.}
\begin{center}
\setlength{\tabcolsep}{0mm}
\begin{tabular}{|c|rrr|rrr|}
\hline
   \hspace{0.007em}  \# \hspace{0.007em} &  \multicolumn{3}{c|}{\hspace{0.02em} \emph{baseline} \hspace{0.07em}} & \multicolumn{3}{c|}{ \emph{+bigram wcNN} }\\ \hline
      1  & \hspace{0.04em} 214/ & 5791  & (3.7\%) \hspace{0.07em} & \hspace{0.02em} 285/ & 5791  & (4.9\%) \\
      2  & \hspace{0.04em} 1225/ & 3200  & (38.3\%) \hspace{0.07em} & \hspace{0.02em} 1243/ & 3200  & (38.8\%) \\
      3  & \hspace{0.04em} 2162/ & 2719  & (79.5\%) \hspace{0.07em} & \hspace{0.02em} 2297/ & 2719  & (84.5\%) \\
      4  & \hspace{0.04em} 3148/ & 3207  & (98.2\%) \hspace{0.07em} & \hspace{0.02em} 3119/ & 3207 & (97.3\%) \\
      5  & \hspace{0.04em} 14602/ & 14602  & (100.0\%) \hspace{0.07em} & \hspace{0.02em} 14602/ & 14602  & (100.0\%) \\ \hline
\end{tabular}
\label{tab:Distri-ArEn}
\end{center}
\end{table}

\section{Conclusion}
\label{sec:syscomNN:conclusion}
In this work we proposed a novel local system voting model (localVote)
which has been trained by a feedforward neural network.
In contrast to the traditional globalVote model, the presented 
localVote model takes the word contents and their combinatorial 
occurrences into account and does not only promote global preferences
for some individual systems. This advantage gives confusion network decoding the option
to prefer other systems at different positions even in the same sentence.
As all words are projected to a continuous space, the neural 
network gives also unseen word sequences a useful probability.
Due to the relatively small neural network training set, we used
word classes in some experiments to tackle the data sparsity problem.

Experiments have been conducted with high quality input systems 
for the BOLT Chinese$\to$English and Arabic$\to$English translation tasks.
Training an additional model by a neural network with word classes yields translation
improvement from up to +0.9 points in \BLEU and -0.5 points in \TER.
We also took word context into account and added the predecessors
of the individual systems to the neural network training which yield additional
small improvement.
We analyzed the translation results and the functionality of the localVote
model. The occurrence distribution shows that words which have been produced by
only few input systems are more likely to be part of the system combination output when using
the proposed model.

\section*{Acknowledgement}
This material is partially based upon work supported by
the DARPA BOLT project under Contract No.
HR0011-12-C-0015. Any opinions, findings and
conclusions or recommendations expressed in this
material are those of the authors and do not necessarily
reflect the views of DARPA.
Further, this paper has received funding from the European Union's Horizon 2020 research and
innovation programme under grant agreement n\textsuperscript{o}~645452 (QT21).

\bibliographystyle{acl}
\bibliography{syscomNN}

\end{document}

%% file: example4.pdf_tex
\begingroup%
  \makeatletter%
  \providecommand\color[2][]{%
    \errmessage{(Inkscape) Color is used for the text in Inkscape, but the package 'color.sty' is not loaded}%
    \renewcommand\color[2][]{}%
  }%
  \providecommand\transparent[1]{%
    \errmessage{(Inkscape) Transparency is used (non-zero) for the text in Inkscape, but the package 'transparent.sty' is not loaded}%
    \renewcommand\transparent[1]{}%
  }%
  \providecommand\rotatebox[2]{#2}%
  \ifx\svgwidth\undefined%
    \setlength{\unitlength}{316.25253906bp}%
    \ifx\svgscale\undefined%
      \relax%
    \else%
      \setlength{\unitlength}{\unitlength * \real{\svgscale}}%
    \fi%
  \else%
    \setlength{\unitlength}{\svgwidth}%
  \fi%
  \global\let\svgwidth\undefined%
  \global\let\svgscale\undefined%
  \makeatother%
  \begin{picture}(1,0.36257649)%
    \put(0,0){\includegraphics[width=\unitlength]{example4.pdf}}%
    \put(0.16583604,0.32590189){\color[rgb]{0,0,0}\makebox(0,0)[lb]{\smash{the}}}%
    \put(0.43665411,0.32590189){\color[rgb]{0,0,0}\makebox(0,0)[lb]{\smash{black}}}%
    \put(0.17058495,0.18493431){\color[rgb]{0,0,0}\makebox(0,0)[lb]{\smash{an}}}%
    \put(0.1829894,0.10193977){\color[rgb]{0,0,0}\makebox(0,0)[lb]{\smash{a}}}%
    \put(0.1829894,0.02714109){\color[rgb]{0,0,0}\makebox(0,0)[lb]{\smash{a}}}%
    \put(0.44322554,0.03183226){\color[rgb]{0,0,0}\makebox(0,0)[lb]{\smash{green}}}%
    \put(0.42701111,0.10663094){\color[rgb]{0,0,0}\makebox(0,0)[lb]{\smash{orange}}}%
    \put(0.4568174,0.17970955){\color[rgb]{0,0,0}\makebox(0,0)[lb]{\smash{red}}}%
    \put(0.74786355,0.32590189){\color[rgb]{0,0,0}\makebox(0,0)[lb]{\smash{cab}}}%
    \put(0.73652852,0.17960951){\color[rgb]{0,0,0}\makebox(0,0)[lb]{\smash{train}}}%
    \put(0.7507316,0.10203982){\color[rgb]{0,0,0}\makebox(0,0)[lb]{\smash{car}}}%
    \put(0.7507316,0.02724114){\color[rgb]{0,0,0}\makebox(0,0)[lb]{\smash{car}}}%
  \end{picture}%
\endgroup%

%% file: example4b.pdf_tex
\begingroup%
  \makeatletter%
  \providecommand\color[2][]{%
    \errmessage{(Inkscape) Color is used for the text in Inkscape, but the package 'color.sty' is not loaded}%
    \renewcommand\color[2][]{}%
  }%
  \providecommand\transparent[1]{%
    \errmessage{(Inkscape) Transparency is used (non-zero) for the text in Inkscape, but the package 'transparent.sty' is not loaded}%
    \renewcommand\transparent[1]{}%
  }%
  \providecommand\rotatebox[2]{#2}%
  \ifx\svgwidth\undefined%
    \setlength{\unitlength}{316.25253906bp}%
    \ifx\svgscale\undefined%
      \relax%
    \else%
      \setlength{\unitlength}{\unitlength * \real{\svgscale}}%
    \fi%
  \else%
    \setlength{\unitlength}{\svgwidth}%
  \fi%
  \global\let\svgwidth\undefined%
  \global\let\svgscale\undefined%
  \makeatother%
  \begin{picture}(1,0.36257649)%
    \put(0,0){\includegraphics[width=\unitlength]{example4b.pdf}}%
    \put(0.16583604,0.32590189){\color[rgb]{0,0,0}\makebox(0,0)[lb]{\smash{the}}}%
    \put(0.17058495,0.18493431){\color[rgb]{0,0,0}\makebox(0,0)[lb]{\smash{an}}}%
    \put(0.1829894,0.10193977){\color[rgb]{0,0,0}\makebox(0,0)[lb]{\smash{a}}}%
    \put(0.1829894,0.02714109){\color[rgb]{0,0,0}\makebox(0,0)[lb]{\smash{a}}}%
    \put(0.43152107,0.17970955){\color[rgb]{0,0,0}\makebox(0,0)[lb]{\smash{UNK}}}%
    \put(0.74786355,0.32590189){\color[rgb]{0,0,0}\makebox(0,0)[lb]{\smash{cab}}}%
    \put(0.73652852,0.17960951){\color[rgb]{0,0,0}\makebox(0,0)[lb]{\smash{train}}}%
    \put(0.7507316,0.10203982){\color[rgb]{0,0,0}\makebox(0,0)[lb]{\smash{car}}}%
    \put(0.7507316,0.02724114){\color[rgb]{0,0,0}\makebox(0,0)[lb]{\smash{car}}}%
  \end{picture}%
\endgroup%

%% file: NN_training_example1.pdf_tex
\begingroup%
  \makeatletter%
  \providecommand\color[2][]{%
    \errmessage{(Inkscape) Color is used for the text in Inkscape, but the package 'color.sty' is not loaded}%
    \renewcommand\color[2][]{}%
  }%
  \providecommand\transparent[1]{%
    \errmessage{(Inkscape) Transparency is used (non-zero) for the text in Inkscape, but the package 'transparent.sty' is not loaded}%
    \renewcommand\transparent[1]{}%
  }%
  \providecommand\rotatebox[2]{#2}%
  \ifx\svgwidth\undefined%
    \setlength{\unitlength}{327.23200684bp}%
    \ifx\svgscale\undefined%
      \relax%
    \else%
      \setlength{\unitlength}{\unitlength * \real{\svgscale}}%
    \fi%
  \else%
    \setlength{\unitlength}{\svgwidth}%
  \fi%
  \global\let\svgwidth\undefined%
  \global\let\svgscale\undefined%
  \makeatother%
  \begin{picture}(1,0.5093766)%
    \put(0,0){\includegraphics[width=\unitlength]{NN_training_example1.pdf}}%
    \put(0.01786872,0.05749919){\color[rgb]{0,0,0}\makebox(0,0)[lb]{\smash{car}}}%
    \put(0.01837946,0.17752406){\color[rgb]{0,0,0}\makebox(0,0)[lb]{\smash{car}}}%
    \put(-0.00051808,0.30265636){\color[rgb]{0,0,0}\makebox(0,0)[lb]{\smash{train}}}%
    \put(0.01327202,0.42778878){\color[rgb]{0,0,0}\makebox(0,0)[lb]{\smash{cab}}}%
    \put(0.84998973,0.44311099){\color[rgb]{0,0,0}\makebox(0,0)[lb]{\smash{$P(w_1|\_)$}}}%
    \put(0.22422255,0.50731394){\color[rgb]{0,0,0}\makebox(0,0)[lb]{\smash{projection
}}}%
    \put(0.50298481,0.43500723){\color[rgb]{0,0,0}\makebox(0,0)[lb]{\smash{hidden
}}}%
    \put(0.26184671,0.45510676){\color[rgb]{0,0,0}\makebox(0,0)[lb]{\smash{layer}}}%
    \put(0.5172639,0.38653836){\color[rgb]{0,0,0}\makebox(0,0)[lb]{\smash{layer}}}%
    \put(0.8498435,0.37858813){\color[rgb]{0,0,0}\makebox(0,0)[lb]{\smash{$P(w_2|\_)$}}}%
    \put(0.85034272,0.31825275){\color[rgb]{0,0,0}\makebox(0,0)[lb]{\smash{$P(w_3|\_)$}}}%
    \put(0.85034272,0.05422004){\color[rgb]{0,0,0}\makebox(0,0)[lb]{\smash{$P(w_n|\_)$}}}%
    \put(0.12216531,0.44992888){\color[rgb]{0,0,0}\makebox(0,0)[lb]{\smash{\tiny{0}}}}%
    \put(0.12216531,0.46741882){\color[rgb]{0,0,0}\makebox(0,0)[lb]{\smash{\tiny{0}}}}%
    \put(0.12216531,0.4324389){\color[rgb]{0,0,0}\makebox(0,0)[lb]{\smash{\tiny{1}}}}%
    \put(0.12216531,0.38609708){\color[rgb]{0,0,0}\makebox(0,0)[lb]{\smash{\tiny{0}}}}%
    \put(0.1270548,0.40658921){\color[rgb]{0,0,0}\makebox(0,0)[b]{\smash{\tiny{.}}}}%
    \put(0.1270548,0.41566375){\color[rgb]{0,0,0}\makebox(0,0)[b]{\smash{\tiny{.}}}}%
    \put(0.1270548,0.42504951){\color[rgb]{0,0,0}\makebox(0,0)[b]{\smash{\tiny{.}}}}%
    \put(0.12310746,0.32504651){\color[rgb]{0,0,0}\makebox(0,0)[lb]{\smash{\tiny{1}}}}%
    \put(0.12310746,0.3425364){\color[rgb]{0,0,0}\makebox(0,0)[lb]{\smash{\tiny{0}}}}%
    \put(0.12310746,0.30755648){\color[rgb]{0,0,0}\makebox(0,0)[lb]{\smash{\tiny{0}}}}%
    \put(0.12310746,0.26121468){\color[rgb]{0,0,0}\makebox(0,0)[lb]{\smash{\tiny{0}}}}%
    \put(0.12799696,0.28170688){\color[rgb]{0,0,0}\makebox(0,0)[b]{\smash{\tiny{.}}}}%
    \put(0.12799696,0.29078133){\color[rgb]{0,0,0}\makebox(0,0)[b]{\smash{\tiny{.}}}}%
    \put(0.12799696,0.30016716){\color[rgb]{0,0,0}\makebox(0,0)[b]{\smash{\tiny{.}}}}%
    \put(0.12266985,0.19941056){\color[rgb]{0,0,0}\makebox(0,0)[lb]{\smash{\tiny{0}}}}%
    \put(0.12266985,0.21690039){\color[rgb]{0,0,0}\makebox(0,0)[lb]{\smash{\tiny{1}}}}%
    \put(0.12266985,0.1819207){\color[rgb]{0,0,0}\makebox(0,0)[lb]{\smash{\tiny{0}}}}%
    \put(0.1221809,0.13557904){\color[rgb]{0,0,0}\makebox(0,0)[lb]{\smash{\tiny{0}}}}%
    \put(0.12755935,0.15607107){\color[rgb]{0,0,0}\makebox(0,0)[b]{\smash{\tiny{.}}}}%
    \put(0.12755935,0.16514565){\color[rgb]{0,0,0}\makebox(0,0)[b]{\smash{\tiny{.}}}}%
    \put(0.12755935,0.17453133){\color[rgb]{0,0,0}\makebox(0,0)[b]{\smash{\tiny{.}}}}%
    \put(0.12203665,0.0762225){\color[rgb]{0,0,0}\makebox(0,0)[lb]{\smash{\tiny{0}}}}%
    \put(0.12203665,0.09371248){\color[rgb]{0,0,0}\makebox(0,0)[lb]{\smash{\tiny{1}}}}%
    \put(0.12203665,0.0587325){\color[rgb]{0,0,0}\makebox(0,0)[lb]{\smash{\tiny{0}}}}%
    \put(0.12203665,0.0123906){\color[rgb]{0,0,0}\makebox(0,0)[lb]{\smash{\tiny{0}}}}%
    \put(0.1269261,0.03288278){\color[rgb]{0,0,0}\makebox(0,0)[b]{\smash{\tiny{.}}}}%
    \put(0.1269261,0.04195733){\color[rgb]{0,0,0}\makebox(0,0)[b]{\smash{\tiny{.}}}}%
    \put(0.1269261,0.0513431){\color[rgb]{0,0,0}\makebox(0,0)[b]{\smash{\tiny{.}}}}%
    \put(0.91913587,0.26166004){\color[rgb]{0,0,0}\makebox(0,0)[lb]{\smash{.}}}%
    \put(0.91913587,0.19062309){\color[rgb]{0,0,0}\makebox(0,0)[lb]{\smash{.}}}%
    \put(0.91913587,0.11958614){\color[rgb]{0,0,0}\makebox(0,0)[lb]{\smash{.}}}%
  \end{picture}%
\endgroup%

%% file: NN_training_example2.pdf_tex
\begingroup%
  \makeatletter%
  \providecommand\color[2][]{%
    \errmessage{(Inkscape) Color is used for the text in Inkscape, but the package 'color.sty' is not loaded}%
    \renewcommand\color[2][]{}%
  }%
  \providecommand\transparent[1]{%
    \errmessage{(Inkscape) Transparency is used (non-zero) for the text in Inkscape, but the package 'transparent.sty' is not loaded}%
    \renewcommand\transparent[1]{}%
  }%
  \providecommand\rotatebox[2]{#2}%
  \ifx\svgwidth\undefined%
    \setlength{\unitlength}{352.25600586bp}%
    \ifx\svgscale\undefined%
      \relax%
    \else%
      \setlength{\unitlength}{\unitlength * \real{\svgscale}}%
    \fi%
  \else%
    \setlength{\unitlength}{\svgwidth}%
  \fi%
  \global\let\svgwidth\undefined%
  \global\let\svgscale\undefined%
  \makeatother%
  \begin{picture}(1,0.47238001)%
    \put(0,0){\includegraphics[width=\unitlength]{NN_training_example2.pdf}}%
    \put(0.02821103,0.06652267){\color[rgb]{0,0,0}\makebox(0,0)[lb]{\smash{green
}}}%
    \put(0.00781126,0.1804485){\color[rgb]{0,0,0}\makebox(0,0)[lb]{\smash{orange
}}}%
    \put(0.06522245,0.2996044){\color[rgb]{0,0,0}\makebox(0,0)[lb]{\smash{red
}}}%
    \put(0.03165756,0.41504933){\color[rgb]{0,0,0}\makebox(0,0)[lb]{\smash{black
}}}%
    \put(0.84702682,0.41082186){\color[rgb]{0,0,0}\makebox(0,0)[lb]{\smash{$P(w_1|\_)$}}}%
    \put(0.26571362,0.47046389){\color[rgb]{0,0,0}\makebox(0,0)[lb]{\smash{projection
}}}%
    \put(0.52013065,0.40783582){\color[rgb]{0,0,0}\makebox(0,0)[lb]{\smash{hidden
}}}%
    \put(0.30066497,0.42196546){\color[rgb]{0,0,0}\makebox(0,0)[lb]{\smash{layer}}}%
    \put(0.54247971,0.36281012){\color[rgb]{0,0,0}\makebox(0,0)[lb]{\smash{layer}}}%
    \put(0.84689098,0.35088252){\color[rgb]{0,0,0}\makebox(0,0)[lb]{\smash{$P(w_2|\_)$}}}%
    \put(0.84735473,0.29483328){\color[rgb]{0,0,0}\makebox(0,0)[lb]{\smash{$P(w_3|\_)$}}}%
    \put(0.84735473,0.04955733){\color[rgb]{0,0,0}\makebox(0,0)[lb]{\smash{$P(w_n|\_)$}}}%
    \put(0.91580304,0.24226088){\color[rgb]{0,0,0}\makebox(0,0)[lb]{\smash{.}}}%
    \put(0.91580304,0.17627034){\color[rgb]{0,0,0}\makebox(0,0)[lb]{\smash{.}}}%
    \put(0.91580304,0.11027986){\color[rgb]{0,0,0}\makebox(0,0)[lb]{\smash{.}}}%
    \put(0.06270741,0.37065868){\color[rgb]{0,0,0}\makebox(0,0)[lb]{\smash{cab}}}%
    \put(0.05297328,0.25541685){\color[rgb]{0,0,0}\makebox(0,0)[lb]{\smash{train}}}%
    \put(0.06806352,0.13514555){\color[rgb]{0,0,0}\makebox(0,0)[lb]{\smash{car}}}%
    \put(0.06806352,0.02153958){\color[rgb]{0,0,0}\makebox(0,0)[lb]{\smash{car}}}%
  \end{picture}%
\endgroup%

%% file: ChEn-BLEUtresholdINK.pdf_tex
\begingroup%
  \makeatletter%
  \providecommand\color[2][]{%
    \errmessage{(Inkscape) Color is used for the text in Inkscape, but the package 'color.sty' is not loaded}%
    \renewcommand\color[2][]{}%
  }%
  \providecommand\transparent[1]{%
    \errmessage{(Inkscape) Transparency is used (non-zero) for the text in Inkscape, but the package 'transparent.sty' is not loaded}%
    \renewcommand\transparent[1]{}%
  }%
  \providecommand\rotatebox[2]{#2}%
  \ifx\svgwidth\undefined%
    \setlength{\unitlength}{364.73601074bp}%
    \ifx\svgscale\undefined%
      \relax%
    \else%
      \setlength{\unitlength}{\unitlength * \real{\svgscale}}%
    \fi%
  \else%
    \setlength{\unitlength}{\svgwidth}%
  \fi%
  \global\let\svgwidth\undefined%
  \global\let\svgscale\undefined%
  \makeatother%
  \begin{picture}(1,0.62767692)%
    \put(0,0){\includegraphics[width=\unitlength]{ChEn-BLEUtresholdINK.pdf}}%
    \put(0.07917405,0.10373633){\makebox(0,0)[lb]{\smash{12}}}%
    \put(0.07917405,0.15966721){\makebox(0,0)[lb]{\smash{13}}}%
    \put(0.07917405,0.2199848){\makebox(0,0)[lb]{\smash{14}}}%
    \put(0.07917405,0.2803024){\makebox(0,0)[lb]{\smash{15}}}%
    \put(0.07917405,0.34062){\makebox(0,0)[lb]{\smash{16}}}%
    \put(0.07917405,0.40093756){\makebox(0,0)[lb]{\smash{17}}}%
    \put(0.07917405,0.46125515){\makebox(0,0)[lb]{\smash{18}}}%
    \put(0.07917405,0.52157275){\makebox(0,0)[lb]{\smash{19}}}%
    \put(0.07917405,0.58189035){\makebox(0,0)[lb]{\smash{20}}}%
    \put(0.13950237,0.06068936){\makebox(0,0)[lb]{\smash{0}}}%
    \put(0.25276068,0.06068936){\makebox(0,0)[lb]{\smash{500}}}%
    \put(0.37752766,0.06068936){\makebox(0,0)[lb]{\smash{1000}}}%
    \put(0.51790094,0.06068936){\makebox(0,0)[lb]{\smash{1500}}}%
    \put(0.65827434,0.06068936){\makebox(0,0)[lb]{\smash{2000}}}%
    \put(0.79878692,0.06068936){\makebox(0,0)[lb]{\smash{2500}}}%
    \put(0.04588507,0.20156037){\rotatebox{90}{\makebox(0,0)[lb]{\smash{$(\TER - \BLEU)/2$}}}}%
    \put(0.41030204,0.00413398){\makebox(0,0)[lb]{\smash{pruning treshold}}}%
    \put(0.67136776,0.54720502){\makebox(0,0)[lb]{\smash{tune (NN)
}}}%
    \put(0.60793144,0.49935857){\color[rgb]{0,0,0}\makebox(0,0)[lb]{\smash{tune (MERT)
}}}%
    \put(0.80244296,0.4479711){\color[rgb]{0,0,0}\makebox(0,0)[lb]{\smash{test}}}%
  \end{picture}%
\endgroup%

%% file: ChEn-wcResults.pdf_tex
\begingroup%
  \makeatletter%
  \providecommand\color[2][]{%
    \errmessage{(Inkscape) Color is used for the text in Inkscape, but the package 'color.sty' is not loaded}%
    \renewcommand\color[2][]{}%
  }%
  \providecommand\transparent[1]{%
    \errmessage{(Inkscape) Transparency is used (non-zero) for the text in Inkscape, but the package 'transparent.sty' is not loaded}%
    \renewcommand\transparent[1]{}%
  }%
  \providecommand\rotatebox[2]{#2}%
  \ifx\svgwidth\undefined%
    \setlength{\unitlength}{331.2bp}%
    \ifx\svgscale\undefined%
      \relax%
    \else%
      \setlength{\unitlength}{\unitlength * \real{\svgscale}}%
    \fi%
  \else%
    \setlength{\unitlength}{\svgwidth}%
  \fi%
  \global\let\svgwidth\undefined%
  \global\let\svgscale\undefined%
  \makeatother%
  \begin{picture}(1,0.6886262)%
    \put(0,0){\includegraphics[width=\unitlength]{ChEn-wcResults.pdf}}%
    \put(0.07580405,0.11163345){\makebox(0,0)[lb]{\smash{21.4}}}%
    \put(0.07580405,0.21308037){\makebox(0,0)[lb]{\smash{21.5}}}%
    \put(0.07580405,0.3193582){\makebox(0,0)[lb]{\smash{21.6}}}%
    \put(0.07580405,0.42563602){\makebox(0,0)[lb]{\smash{21.7}}}%
    \put(0.07580405,0.53192564){\makebox(0,0)[lb]{\smash{21.8}}}%
    \put(0.07580405,0.63337254){\makebox(0,0)[lb]{\smash{21.9}}}%
    \put(0.21724999,0.06422771){\makebox(0,0)[lb]{\smash{500}}}%
    \put(0.34951982,0.05939679){\makebox(0,0)[lb]{\smash{1000}}}%
    \put(0.49897634,0.05939679){\makebox(0,0)[lb]{\smash{1500}}}%
    \put(0.64843287,0.05939679){\makebox(0,0)[lb]{\smash{2000}}}%
    \put(0.79803087,0.05939679){\makebox(0,0)[lb]{\smash{2500}}}%
    \put(0.6762099,0.19201822){\makebox(0,0)[lb]{\smash{\emph{baseline}}}}%
    \put(0.50220396,0.13767039){\makebox(0,0)[lb]{\smash{\emph{+unigram wcNN}}}}%
    \put(0.0442614,0.23993632){\rotatebox{90}{\makebox(0,0)[lb]{\smash{$(\TER - \BLEU)/2$}}}}%
    \put(0.44826395,0.00542532){\makebox(0,0)[lb]{\smash{word class size}}}%
  \end{picture}%
\endgroup%

%% file: ArEn-BLEUtresholdINK.pdf_tex
\begingroup%
  \makeatletter%
  \providecommand\color[2][]{%
    \errmessage{(Inkscape) Color is used for the text in Inkscape, but the package 'color.sty' is not loaded}%
    \renewcommand\color[2][]{}%
  }%
  \providecommand\transparent[1]{%
    \errmessage{(Inkscape) Transparency is used (non-zero) for the text in Inkscape, but the package 'transparent.sty' is not loaded}%
    \renewcommand\transparent[1]{}%
  }%
  \providecommand\rotatebox[2]{#2}%
  \ifx\svgwidth\undefined%
    \setlength{\unitlength}{360bp}%
    \ifx\svgscale\undefined%
      \relax%
    \else%
      \setlength{\unitlength}{\unitlength * \real{\svgscale}}%
    \fi%
  \else%
    \setlength{\unitlength}{\svgwidth}%
  \fi%
  \global\let\svgwidth\undefined%
  \global\let\svgscale\undefined%
  \makeatother%
  \begin{picture}(1,0.61555556)%
    \put(0,0){\includegraphics[width=\unitlength]{ArEn-BLEUtresholdINK.pdf}}%
    \put(0.09008264,0.10027778){\makebox(0,0)[lb]{\smash{4}}}%
    \put(0.09008264,0.17735461){\makebox(0,0)[lb]{\smash{6}}}%
    \put(0.09008264,0.25874567){\makebox(0,0)[lb]{\smash{8}}}%
    \put(0.07594418,0.34027779){\makebox(0,0)[lb]{\smash{10}}}%
    \put(0.07594418,0.42179901){\makebox(0,0)[lb]{\smash{12}}}%
    \put(0.07594418,0.50319007){\makebox(0,0)[lb]{\smash{14}}}%
    \put(0.07594418,0.58472219){\makebox(0,0)[lb]{\smash{16}}}%
    \put(0.13706615,0.06110894){\makebox(0,0)[lb]{\smash{0}}}%
    \put(0.24737,0.06110894){\makebox(0,0)[lb]{\smash{500}}}%
    \put(0.3782228,0.06110894){\makebox(0,0)[lb]{\smash{1000}}}%
    \put(0.52044277,0.06110894){\makebox(0,0)[lb]{\smash{1500}}}%
    \put(0.66266286,0.06110894){\makebox(0,0)[lb]{\smash{2000}}}%
    \put(0.80502397,0.06110894){\makebox(0,0)[lb]{\smash{2500}}}%
    \put(0.41206306,-0.00380955){\makebox(0,0)[lb]{\smash{pruning treshold}}}%
    \put(0.04458333,0.19806064){\rotatebox{90}{\makebox(0,0)[lb]{\smash{$(\TER - \BLEU)/2$}}}}%
    \put(0.67929032,0.55641239){\makebox(0,0)[lb]{\smash{tune (NN)
}}}%
    \put(0.61587274,0.50793648){\color[rgb]{0,0,0}\makebox(0,0)[lb]{\smash{tune (MERT)}}}%
    \put(0.81333381,0.46031744){\color[rgb]{0,0,0}\makebox(0,0)[lb]{\smash{test}}}%
  \end{picture}%
\endgroup%

%% file: ArEn-wcResults.pdf_tex
\begingroup%
  \makeatletter%
  \providecommand\color[2][]{%
    \errmessage{(Inkscape) Color is used for the text in Inkscape, but the package 'color.sty' is not loaded}%
    \renewcommand\color[2][]{}%
  }%
  \providecommand\transparent[1]{%
    \errmessage{(Inkscape) Transparency is used (non-zero) for the text in Inkscape, but the package 'transparent.sty' is not loaded}%
    \renewcommand\transparent[1]{}%
  }%
  \providecommand\rotatebox[2]{#2}%
  \ifx\svgwidth\undefined%
    \setlength{\unitlength}{350.4bp}%
    \ifx\svgscale\undefined%
      \relax%
    \else%
      \setlength{\unitlength}{\unitlength * \real{\svgscale}}%
    \fi%
  \else%
    \setlength{\unitlength}{\svgwidth}%
  \fi%
  \global\let\svgwidth\undefined%
  \global\let\svgscale\undefined%
  \makeatother%
  \begin{picture}(1,0.64546441)%
    \put(0,0){\includegraphics[width=\unitlength]{ArEn-wcResults.pdf}}%
    \put(0.10444108,0.09552148){\makebox(0,0)[lb]{\smash{9.8}}}%
    \put(0.11169841,0.21652606){\makebox(0,0)[lb]{\smash{10}}}%
    \put(0.08991527,0.34209683){\makebox(0,0)[lb]{\smash{10.2}}}%
    \put(0.08991527,0.46766757){\makebox(0,0)[lb]{\smash{10.4}}}%
    \put(0.08991527,0.59323834){\makebox(0,0)[lb]{\smash{10.6}}}%
    \put(0.21576921,0.05984573){\makebox(0,0)[lb]{\smash{500}}}%
    \put(0.41583517,0.05984573){\makebox(0,0)[lb]{\smash{1500}}}%
    \put(0.62788078,0.05984573){\makebox(0,0)[lb]{\smash{2500}}}%
    \put(0.83978146,0.05984573){\makebox(0,0)[lb]{\smash{3500}}}%
    \put(0.70082606,0.49863982){\makebox(0,0)[lb]{\smash{\emph{baseline}}}}%
    \put(0.52565107,0.44726995){\makebox(0,0)[lb]{\smash{\emph{+unigram wcNN}}}}%
    \put(0.04865967,0.21024446){\rotatebox{90}{\makebox(0,0)[lb]{\smash{$(\TER - \BLEU)/2$}}}}%
    \put(0.4500355,0.00056185){\makebox(0,0)[lb]{\smash{word class size}}}%
  \end{picture}%
\endgroup%